\documentclass{article}

\usepackage{arxiv}

\usepackage[utf8]{inputenc} 
\usepackage[T1]{fontenc}    
\usepackage{hyperref}       
\usepackage{url}            
\usepackage{booktabs}       
\usepackage{amsfonts}       
\usepackage{nicefrac}       
\usepackage{microtype}      
\usepackage{xcolor}         

\title{SMLT-MUGC: Small, Medium, and Large Texts - Machine versus User Generated Content Detection and Comparison}

\author{
    Anjali Rawal \\
  Northeastern University\\
  Seattle WA 98109 \\
  \texttt{rawal.an@northeastern.edu} \\
\And
    Hui Wang \\
  Northeastern University\\
  Seattle WA 98109 \\
  \texttt{wang.hui7@northeastern.edu} \\
\And
  Youjia Zheng\\
  Northeastern University\\
  Seattle WA 98109 \\
  \texttt{wang.hui7@northeastern.edu} \\
\And
Yu-Hsuan Lin\\
  Northeastern University\\
  Seattle WA 98109 \\
  \texttt{lin.yu-h@northeastern.edu} \\
  \And
 Shanu Sushmita \\
 Northeastern University\\
  Seattle WA 98109 \\
  \texttt{s.sushmita@northeastern.edu} \\
}

\begin{document}

\maketitle
\begin{abstract}

Large language models (LLMs) have attracted significant attention in both academia and industry due to their impressive ability to mimic human language. Accurately identifying texts generated by LLMs is essential for understanding their full capabilities and mitigating potential serious consequences. This paper presents a comprehensive analysis across datasets of varying text lengths—\textit{small, medium, and large}. We compare the performance of different machine learning algorithms on four extensive datasets: \textit{small} (tweets from Election, FIFA, and Game of Thrones), \textit{medium} (Wikipedia introductions and PubMed abstracts), and \textit{large} (OpenAI web text dataset). Our results indicate that large language models (LLMs) with very large parameters (such as the XL-1542 variant of the GPT-2 model, which has 1542 million parameters) were harder ($\approx 74\%$) to detect using traditional machine learning methods. However, detecting texts of varying lengths (small, medium, and large) from LLMs with smaller parameters (762 million or less) can be done with very high accuracy ($\approx 96\%$ and above). Additionally, we examine the characteristics of human and machine-generated texts across multiple dimensions, including linguistics, personality, sentiment, bias, and morality. Our findings indicate that machine-generated texts generally have higher readability and closely mimic human moral judgments. However, they exhibit notable differences in personality traits compared to human-generated texts. SVM and Voting Classifier (VC) models consistently achieve high performance across most datasets, while Decision Tree (DT) models show the lowest performance. We also find that model performance drops when dealing with rephrased texts, particularly for shorter texts like tweets. This study underscores the challenges and importance of detecting LLM-generated texts and suggests directions for future research to improve detection methods and understand the nuanced capabilities of LLMs.
\end{abstract}

\section{Introduction}

 Modern large language models are becoming increasingly powerful and capable of generating realistic and convincing content. These models can also target users with highly personalized recommendations. Consequently, they have the potential to create and propagate harmful or misleading content, such as fake news or hate speech, whether intentionally or unintentionally \cite{Emsley:2023,deshpande:2023,Taloni:2023}. The proliferation of content generated and distributed by these models has amplified the threat and impact of such harmful activities more than ever before \cite{solaiman:2019,Zellers:2019,deshpande:2023,Taloni:2023, Else:2023}.
As AI-generated texts increasingly blend seamlessly with human-written content, the demand for more effective methods to detect misleading information produced by AI grows. Therefore, there is a growing interest in investigating the impact and detection mechanisms for machine-generated content \cite{tang2023,Islam:2023,yucheat:2023,guo2023,Lancet:2024,Tang:2024}. Accurately detecting machine-generated content is essential to understand their full capabilities while minimizing the potential for any serious consequences. In prior detection efforts, researchers predominantly utilized pre-trained models like RoBERTa, GPT-2, GROVER, and GLTR \cite{Jawahar:2020}, and more recently, the effectiveness of conventional methods such as logistic regression, SVM, and others is also being explored \cite{xie:2024,Islam:2023}. Overall, supervised detection, zero-shot detection, retrieval-based detection, and watermarking-based detection are being used to distinguish machine-generated text from human text \cite{Yang:2023,abdali:2024}. Despite growing interest and several published studies, a consolidated comparison across different domains, text lengths, and characteristics is still missing.

This paper aims to contribute towards that goal by presenting three detailed analyses on datasets comprising varying text lengths—\textit{small, medium, and large}. First, it compares the performance of different machine learning algorithms on four datasets: \textit{small} (tweets from Election, FIFA, and Game of Thrones), \textit{medium} (Wikipedia introductions and PubMed abstracts), and \textit{large} (OpenAI web text dataset). Second, it compares the characteristics of human and machine-generated writing across multiple dimensions, including linguistics, personality, sentiment, bias, and morality. Third, it compares the algorithm performances on rephrased texts. Our analysis revealed several key findings: Machine-generated texts are generally more accessible and require a lower level of education to understand compared to human-generated texts. While there are some differences in sentiment scores, especially in the weak negative and weak positive metrics, the scores are generally close across most datasets, indicating that machine-generated texts can closely mimic the sentiment characteristics of human-generated texts. Machine-generated texts show notable differences in personality metrics compared to human-generated texts across different datasets. SVM and Voting Classifier (VC) consistently achieve high performance across most datasets, particularly excelling in the Wiki and Abstract datasets. In contrast, Decision Tree (DT) models generally exhibited the lowest performance across all datasets. Finally, the performance of models drops when dealing with rephrased data. This drop is more pronounced for short texts like tweets compared to medium-length texts like PubMed abstracts.

The rest of the paper is organized as follows: In Section \ref{related}, we discuss the related background. The datasets used in this study are described in Section \ref{dataset}, whereas their linguistic, personality, and emotional characteristics of human and machine-generated data are presented in Section \ref{char}. The performance results of the machine learning algorithms for the task of detection are discussed in Section \ref{result}. Section \ref{rephrase} presents the implications of re-phrasing techniques on the model performances. Finally, in Section~\ref{conc}, we conclude with our overall findings.

\section{Related Work}
\label{related}

Large language models (LLMs), such as the Generative Pre-trained Transformer (GPT) models developed by OpenAI, have demonstrated remarkable capabilities in generating human-like text across various domains \cite{radford:2018}. These models are trained on extensive datasets and can produce coherent and contextually relevant text in response to prompts. However, alongside their transformative potential, LLMs pose significant challenges, particularly regarding the identification and mitigation of potential misuse or harmful outputs. The number of users and applications for these models is growing at an unprecedented rate. Most recently, ChatGPT reported having 180 million users \cite{Duarte:2024}. This rapid growth highlights the profound impact of this technology on individuals and society \cite{Else:2023, Varanasi:2023, Rooney:2023, Oravec:2023, Yang:2023}.

The influence of LLMs in education has raised substantial concerns. While their convenience is notable, the risk of providing swift answers poses a threat to the development of critical thinking and problem-solving skills, which are essential for academic and lifelong success. Additionally, there's concern about academic honesty, as students may be tempted to use these tools inappropriately. In response, New York City Public Schools have prohibited the use of ChatGPT \cite{Rooney:2023}. Although the impact of LLMs on education is considerable, it is crucial to extend this discussion to other domains as well. In journalism, for example, the emergence of AI-generated "deepfake" news articles could jeopardize the credibility of news outlets and misinform the public. In the legal sector, the potential misuse of LLMs could have repercussions on the justice system, affecting processes from contract generation to litigation. Furthermore, in cybersecurity, LLMs could be weaponized to craft more convincing phishing emails or execute social engineering attacks. Zellers et al. \cite{Zellers:2019} found that Grover (Generating aRticles by Only Viewing mEtadata Records) can rewrite propaganda articles, with humans rating the rewritten versions as more trustworthy. LLMs when assigned a persona can show increase in toxicity \cite{deshpande:2023}. Depending on the persona assigned to ChatGPT, its toxicity can increase up to 6 times, with outputs engaging in incorrect stereotypes, harmful dialogue, and hurtful opinions. 

Detecting text generated by LLMs is crucial for various reasons, including combating misinformation, protecting against online abuse, and ensuring the ethical use of AI technologies. Recent research has focused on developing methodologies for distinguishing between text generated by LLMs and human-authored content. Approaches range from linguistic analysis and stylometric features to leveraging artifacts specific to model architectures \cite{sap:2020}. One common strategy involves analyzing linguistic patterns and semantic coherence in generated text. LLMs often exhibit characteristic biases, repetition, or semantic inconsistencies that differ from human-written text \cite{bender:2021}. Additionally, researchers have explored the use of stylometric features, such as writing style, vocabulary usage, and syntactic structures, to differentiate between LLM-generated text and human-written text \cite{arnold:2022}. 

Advancements in adversarial testing frameworks have enabled the evaluation of LLM robustness against detection techniques. Adversarial examples, crafted to evade detection mechanisms, provide insights into the vulnerabilities of detection models and inform the development of more robust detection strategies \cite{gururangan:2020}. Jonathan \cite{Pan:2023} illustrated cybersecurity issues raised by machine-generated texts, the importance of distinguishing machine-generated texts from human-generated texts, and how to use a model to classify machine-generated texts. Using the Foundation Model framework, they trained their own model using human-generated texts with self-supervised learning and then fine-tuned the model using a transfer learning approach on a small set of machine-generated texts. Although the results are promising, further improvements are required to accommodate the growing LLM model.

In the work done by Mohamed et al. \cite{Abdalla:2023}, they assessed the difficulty and importance of identifying genuine research in academic writing and scientific publication. They developed datasets of human-written and machine-generated scientific papers using various generative AI and experimented with classifiers to build a model to detect authorship. The classifiers were then assessed in terms of generalization capabilities and explainability. Chong et al. \cite{Chong:2023} discussed the identification of critical linguistic, emoji, and sentiment characteristics to distinguish between machine-generated and human-generated posts on Twitter through a classifier. Their model includes BERT embeddings with semantic, emoji, sentiment, and linguistic features, and a multi-layer perceptron is used to incorporate additional features beyond BERT. It was evaluated using logistic regression, support vector machines, and random forest. They found apparent differences between machine-generated texts and human-generated texts according to these features, achieving accuracy scores of 88.3\%, with F1 scores ranging between 88.1\% and 88.3\%.

Xie et al. \cite{xie:2024} tested traditional machine learning methods like logistic regression, support vector machines, decision tree, k-nearest neighbor, random forest, AdaBoost, bagging classifier, gradient boosting, multi-layer perceptron, long short-term memory, etc., for the task of machine-generated text detection. While \cite{Islam:2023} observed that extremely randomized trees exhibited the highest F1-score at 76\%, \cite{xie:2024} reported very high accuracy (above 95\%) on abstract, PubMed, and poetry datasets. Gehrmann et al. \cite{Gehrmann:2019} used a statistical approach to detect differences in patterns between human and model-generated texts. Assuming that AI systems tend to overgenerate from a limited subset of the true distribution of natural language, they developed a tool called GLTR (Giant Language model Test Room) which assesses each word in terms of its probability, absolute rank, and entropy of its prediction. While GLTR significantly improved the detection rate (from 54\% to 72\%) of generated text, the tool relies heavily on the assumption that models have a tendency for biased sampling. Overall, the detection of text generated by LLMs is a multidisciplinary endeavor, drawing on insights from natural language processing, machine learning, and computational linguistics. Addressing the challenges associated with large language model detection is essential for fostering trust, transparency, and the responsible deployment of AI technologies in various domains.

\section{Dataset}
\label{dataset}

In this section, we describe the four datasets of varying lengths used in our study: (1) OpenAI GPT-2 Webtext dataset; (2) Wikipedia introductions; (3) Pubmed abstracts; and (4) a Twitter dataset. Before conducting the statistical analysis and modeling, we adhered to standard data cleaning procedures. These included removing missing values, stopwords (except for the OpenAI and Twitter datasets), non-English text, excess whitespaces, special characters, and isolated single digits. The decision not to remove stopwords from the OpenAI and Twitter datasets was based on observed performance degradation in detection tasks when stopwords were excluded. Each dataset contains a 'label' column that indicates whether the text was authored by a machine or a human ( $1= machine$ and $0=human$). Overview of the four dataset description is provided in the Table \ref{tab:mean_SD}. The table provides insights into the differences between machine-generated and human-generated texts across various datasets in terms of token count and vocabulary size. Human texts generally have larger vocabularies, while the length of the texts varies depending on the dataset. The machine-generated texts tend to be more consistent in length compared to human texts, which show more variability. For our experiments,  we used the University HPC Cluster which is a high-powered, multi-node, parallel computing system designed to support large datasets. More specifically, we used jupyter labs on the RC using short partition and 6 CPUs, and max of 8gb memory. Details of each dataset is described next. 

\begin{table}[ht]
   \caption{Mean distribution of tokens/words per text (mean and SD), dataset size (n) and their corresponding columns.}
    \label{tab:mean_SD}
    \small
    \begin{tabular}{ccccc}
     \toprule 
       \textbf{Dataset} &	\textbf{Tokens-Machine} & 	\textbf{Tokens-Human} & \textbf{Total (n)} & \textbf{Vocab (human/Machine)}\\ \hline 
OpenAI- GPT2 &	481.87 (277.43) & 440.94 (275.87) & 500,000 & 2,437,481/3,738,722  \\  
Wiki Intro	& 50.60 (21.06) & 82.32 (18.54)	 &300,000 &  465,420 / 213,963  \\ 
Pubmed & 105 (41.64)& 132.73 (80.66) & 29,844 & 88,317/24,225\\ 	
Twitter &	22.77(4.87)&	 28.48 (15.94)& 59,999& 32,730/7,735\\ 
\bottomrule
 \end{tabular}
\end{table}

\paragraph{Open AI GPT-2 Output Dataset (Long Text Length):}

The OpenAI GPT-2 Output Dataset \cite{gpt2outputdataset} consists of text generated by the GPT-2 model, a transformer-based language model developed by OpenAI. This dataset was created to study GPT-2's outputs and provide a resource for researchers investigating language generation, model evaluation, and detection of machine-generated text. Key characteristics include: (i) a variety of prompts that vary in topic and style, reflecting the model's versatility; (ii) text generated by different sizes of GPT-2, from 117M to 1.5B parameters, allowing for comparison across model capacities. The dataset comprises 250K documents from the WebText test set, including 250K random samples and 250K samples generated using Top-K truncation of 40. Table \ref{tab:mean_SD} provides initial statistics: the average number of tokens in machine-generated text is 481.87, slightly higher than the 440.94 in human-generated text, with both showing high variability (standard deviations around 277). Machine-generated text has a significantly larger vocabulary (3,738,722) compared to human-generated text (2,437,481). 

\paragraph{GPT Wiki Intro (Medium Text Length):}
GPT Wki Intro \cite{aaditya_bhat_2023} is an open source dataset for training models to classify human-written vs. GPT/ChatGPT generated text. This dataset comprises Wikipedia introductions and GPT (Curie) generated introductions for 150K topics. Prompt used for generating text: A 200-word Wikipedia-style introduction on \texttt{\{title\}}: \texttt{\{starter\_text\}} where \texttt{\{title\}} is the title for the Wikipedia page, and \texttt{\{starter\_text\}} is the first seven words of the Wikipedia introduction. From the Table \ref{tab:mean_SD}, it can be seen that the Machine-generated texts are shorter on average (50.60 tokens) compared to human-generated texts (82.32 tokens). The standard deviations are relatively low, indicating consistent lengths. Human-generated text has a larger vocabulary (465,420) than machine-generated text (213,963). In total, there are 300,000 wikipedia text samples.

\paragraph{Pubmed Abstract (Medium Text Length):}
For the purpose of investigating the technical writing styles (dataset), we utilized the Pubmed Python library\footnote{https://pypi.org/project/pymed/} to extract abstracts from the vast repository of scientific literature available on Pubmed. Our focus was particularly directed towards abstracts and their corresponding titles related to the topic of HIV and Dementia. To balance our dataset and to generate equivalent machine-generated content, we used the following approach. Each human-generated abstract's title served as a prompt for ChatGPT-3, enabling us to generate machine-equivalent abstracts. Through this method, we successfully generated approximately 30K abstracts for the purpose of this study. Overall, machine-generated texts in pubmed dataset have an average of 105 tokens, while human-generated texts are longer with an average of 132.73 tokens (See Table \ref{tab:mean_SD}. Additionally, human-generated texts show more variability (standard deviation of 80.66) compared to machine-generated texts (standard deviation of 41.64). The vocabulary size for human-generated text (88,317) is larger than for machine-generated text (24,225).

\paragraph{Tweets (Short Text Length):}
To evaluate the detection task on short text samples, tweets from three topics were utilized: US Election 2020\footnote{https://www.kaggle.com/datasets/manchunhui/us-election-2020-tweets}, FIFA World Cup 2022\footnote{https://www.kaggle.com/datasets/tirendazacademy/fifa-world-cup-2022-tweets}, and Game of Thrones Season 8\footnote{https://www.kaggle.com/datasets/monogenea/game-of-thrones-twitter}. These datasets represented the 'Human' dataset. Machine-generated tweets were produced using the OpenAI library with the 'gpt-3.5-turbo' model. The prompt used for generating tweets was: "Generate \{max\_tweets\_per\_request\} tweets using the following prompt: You are a devoted supporter of Biden. Write an enthusiastic twitter post to show your support for him in the 2020 election. Do not exceed \{character\_limit\} characters per tweet. Do not include line breaks within a tweet." For the US Election Twitter dataset, tweets were balanced to demonstrate support for both the Democratic and Republican parties. This approach aimed to ensure equal representation and a comprehensive analysis. The diverse topics create a robust benchmark for comparing human and machine-generated texts, introducing variety into the classification task. Ten thousand tweets were randomly sampled from the 'Human' dataset, totaling 30K human tweets. Using gpt-3.5, another 30K machine tweets were generated, resulting in 60K tweets in total. Data statistics are provided in Table \ref{tab:mean_SD}. On average, human-generated tweets contain more tokens (28.48) compared to machine-generated tweets (22.77). This suggests that human tweets tend to be longer. Human-generated tweets exhibit greater variability in length (standard deviation of 15.94) compared to machine-generated tweets (standard deviation of 4.87). he human-generated text has a significantly larger vocabulary (32,730 unique words) compared to the machine-generated text (7,735 unique words). This indicates that human-generated tweets use a more diverse set of words than machine-generated tweets.

\section{Characteristic Analysis}
\label{char}
To understand the characteristics of human and machine-generated data, we analyzed them across various linguistic, personality, and emotional dimensions. The rationale for including linguistic and emotional features is that individuals express themselves differently, using distinct words, phrases, and emotions (e.g., anger, joy). In contrast, machine-generated content tends to lack strong emotions or specific personality traits. Research has shown that machine-generated writing is typically polite, devoid of specific details, uses sophisticated and unusual vocabulary, is impersonal, and generally does not convey emotions \cite{mitrović2023chatgpt}. Additionally, while machine-generated medical content is grammatically flawless and human-like, its linguistic characteristics differ from those written by human experts \cite{liao2023differentiate}. The difference between the human and machine data scores were compared for significance using t-test at $\alpha =0.05$. 

\paragraph{Readability:}
To analyze the complexity and readability of the poems, essays and abstracts, we utilized four well known readability metrics \cite{McCLURE:1987} -- Gunning Fog Index\footnote{https://en.wikipedia.org/wiki/Gunning\_fog\_index}, SMOG Index\footnote{https://readable.com/readability/smog-index/}, Dale-Chall Readability Score\footnote{https://readabilityformulas.com/word-lists/the-dale-chall-word-list-for-readability-formulas/}, Flesch Reading Ease Score\footnote{https://yoast.com/flesch-reading-ease-score/} and coleman liau index \cite{horne2019robust}. 
Research pertaining to Large Language Models (LLMs) and their linguistic capabilities has been of significant interest to researchers in the past \cite{beguš:2023}. However, there has been a relatively limited exploration from the perspective of readability scores. 
The scores for both machine and human data are provided in the Table \ref{tab:read}. The results in the table suggest that machine-generated texts tend to be simpler and more accessible than human-generated texts, especially in the Wiki and Tweets datasets. However, in the OpenAI and PubMed datasets, the results are mixed, with machine-generated texts sometimes being more complex or slightly easier to read, depending on the metric. This indicates that while machine-generated texts can mimic human writing closely, they often result in more straightforward and less nuanced language.

\begin{table}
\caption{Average readability scores for human and machine generated texts.}
\label{tab:read}
\small
\begin{tabular}{ccccccccc}
     \toprule
 & \multicolumn{2}{|c|}{OpenAI} & \multicolumn{2}{|c|}{Wiki} & \multicolumn{2}{|c|}{Pubmed} & \multicolumn{2}{|c|}{Tweets} \\ 
 \cmidrule(r){1-9}
 \textbf{Metric}& \textbf{Human} & \textbf{Machine} & \textbf{Human} & \textbf{Machine} & \textbf{Human} & \textbf{Machine} & \textbf{Human} & \textbf{Machine}\\ 
\midrule 
Gunning Fog Index & 11.62 & 12.50 & 12.89&10.41& 16.82& 15.06& 14.72& 12.05\\
SMOG Index & 11.26 & 11.42 &12.76&11.08 &17.16 & 14.93& 11.75& 10.23\\
Dale-Chall Readability  & 9.23 & 9.38  &10.14& 9.54& 10.86&11.40&10.66& 9.55\\
Flesch Reading Ease Score & 59.35 & 58.90  & 53.19 & 60.97& 19.99& 29.95&56.62& 68.86\\
Coleman Liau Index &  10.74& 10.66 & 11.69 & 10.45& 18.40& 16.42&11.75& 9.06\\
\bottomrule
\end{tabular}
    
\end{table}

\paragraph{Bias:}
This set of features encapsulates the general bias and subjectivity present in the text. It draws heavily from the research of Recasens et al. [46] in identifying biased language. This feature set encompasses various linguistic indicators, such as hedges, factives, assertives, implicatives, and opinion words. Additionally, it incorporates the count of biased words according to a lexicon of biased terms referenced. Results for bias features are shown in the Table \ref{tab:bias}. The results in the table indicates that machine-generated texts tend to exhibit slightly higher scores in several bias metrics compared to human-generated texts. This is particularly notable in the "Bias words" and "Factives" metrics across the Wiki and Pubmed datasets, suggesting that machine-generated texts may contain more biased language and assertive statements. However, for "Hedges" and "Implicatives," the scores vary, with some datasets showing higher scores for human-generated texts and others for machine-generated texts. Overall, while the differences are generally small, machine-generated texts exhibit a slight tendency towards higher bias across most metrics.

\begin{table}
\caption{Bias metric comparisons between human and machine generated data. Here, mean score for dataset are reported. }
\label{tab:bias}
\small
\begin{tabular}{ccccccccc}
     \toprule
 & \multicolumn{2}{|c|}{OpenAI} & \multicolumn{2}{|c|}{Wiki} & \multicolumn{2}{|c|}{Pubmed} & \multicolumn{2}{|c|}{Tweets} \\ 
 \cmidrule(r){1-9}
 \textbf{Metric}& \textbf{Human} & \textbf{Machine} & \textbf{Human} & \textbf{Machine} & \textbf{Human} & \textbf{Machine} & \textbf{Human} & \textbf{Machine}\\ 
\midrule 
Bias words & 0.0781 & 0.0744 &0.2288 &0.2495 &0.0599 &0.0727 &0.1246& 0.1330\\
Assertatives & 0.0044& 0.0043 &0.0105 &0.0105&0.0020&0.0029&0.0097& 0.0100\\
Factives & 0.0025& 0.0026 &0.0051&0.0058 &0.0024 &0.0034 &0.0062& 0.0074\\
Hedges & 0.0117 & 0.0116 &0.0107&0.0093 &0.0077 &0.0068 &0.0106& 0.0119\\
Implicatives & 0.0059 & 0.0054 &0.0046 &0.0039 &0.0049 &0.0060 &0.0111& 0.0180\\
\bottomrule
\end{tabular}
\end{table}

\paragraph{Moral:}
This feature set draws from Moral Foundation Theory \cite{graham:2013} and lexicons referenced in \cite{li:2023}. Despite its application in prior research, its efficacy in the news context or its ability to capture significant signals has not been demonstrated. We include this feature group for comprehensive analysis purposes. In total, this group comprises 10 features as shown in Table \ref{tab:moral}. The Fairness score are almost identical across human and machine-generated texts for OpenAI and Tweets datasets. In the Wiki dataset, human texts score slightly higher. Whereas, Cheating was only present in the Wiki dataset, where machine-generated texts have a higher score (0.0040) compared to human texts (0.0035). Very similar `Harm' scores were observed across human and machine-generated texts for all datasets. Machine texts in Wiki and Pubmed datasets show slightly higher scores. Overall, the results indicates that the morality metrics between human and machine-generated texts are generally similar, with some variations. Machine-generated texts tend to have slightly higher scores in certain metrics like "Cheating" and "Harm" in specific datasets. Overall, the differences are minor, suggesting that both types of texts exhibit similar moral considerations across the datasets. This similarity highlights the capability of machine-generated texts to closely mimic human moral judgments in content.  

\begin{table}
\small
\caption{Morality comparisons between human and machine generated data. Here, mean for dataset are reported. }
\label{tab:moral}
\begin{tabular}{ccccccccc}
     \toprule
 & \multicolumn{2}{|c|}{OpenAI} & \multicolumn{2}{|c|}{Wiki} & \multicolumn{2}{|c|}{Pubmed} & \multicolumn{2}{|c|}{Tweets} \\ 
 \cmidrule(r){1-9}
 \textbf{Metric}& \textbf{Human} & \textbf{Machine} & \textbf{Human} & \textbf{Machine} & \textbf{Human} & \textbf{Machine} & \textbf{Human} & \textbf{Machine}\\ 
\midrule 
 Harm & 0.0019 & 0.0019 &0.0035&0.0040 &0.0012&.0014 &0.0019& 0.0019\\
Fairness & 0.0003 & 0.0003 &0.0003&0.0002 &0.0001 &0.000 &0.0003& 0.0003\\
Cheating & 0.000 & 0.000 &0.0035 &0.0040&0.0001&0.000&0.000& 0.000\\
Loyalty & 0.0017 & 0.0017 &0.0058 &0.0061&0.0027&0.0020&0.0010& 0.0007\\
Betrayal & 0.0002  & 0.0002 &0.0004&0.0003&0.000&0.000&0.0002& 0.0001\\
Authority & 0.0025 & 0.0025 &0.0063 &0.0048&0.002&0.001&0.0012& 0.0010\\
Subversion & 0.0002 & 0.0002 &0.0003&0.0003&0.000&0.000&0.0003& 0.0002\\
Purity & 0.0003 & 0.0003 &0.0015&0.0013&0.000&0.000&0.0002& 0.0003\\
Degradation & 0.0002 & 0.0003 &0.0003 &0.0003&0.000&0.00003&0.0005& 0.0005\\
Morality General & 0.0016 & 0.0018 &0.0013&0.0013&0.0014&0.0010&0.0042& 0.0034\\
\bottomrule
\end{tabular}
    
\end{table}

\paragraph{Sentiment:}
To investigate the expression style of human and machine writing we computed various sentiment features from Liu et. al \cite{Liu:2005} using python NELA Feature extractor\footnote{https://pypi.org/project/nela-features/}. Results are provided in Table \ref{tab:sentiment}. The table provides insights into how sentiment scores vary between human and machine-generated texts. For instance, scores are similar for OpenAI human and machine-generated texts (0.0250), and machine-generated texts have slightly higher weak positive scores in the Wiki, Pubmed, and Tweets datasets compared to human-generated texts. While there are some differences in sentiment scores, especially in the weak negative and weak positive metrics, the scores are generally close across most datasets, indicating that machine-generated texts can closely mimic the sentiment characteristics of human-generated texts. However, certain nuances, such as higher weak negative sentiment in machine texts for the OpenAI dataset, highlight areas where machine-generated content differs from human-generated content. 

\begin{table}
\caption{Sentiment comparisons between human and machine generated data. Here, average scores for datasets are reported. }
\label{tab:sentiment}
\small
\begin{tabular}{ccccccccc}
     \toprule
 & \multicolumn{2}{|c|}{OpenAI} & \multicolumn{2}{|c|}{Wiki} & \multicolumn{2}{|c|}{Pubmed} & \multicolumn{2}{|c|}{Tweets} \\ 
 \cmidrule(r){1-9}
 \textbf{Metric}& \textbf{Human} & \textbf{Machine} & \textbf{Human} & \textbf{Machine} & \textbf{Human} & \textbf{Machine} & \textbf{Human} & \textbf{Machine}\\ 
\midrule 
Weak Negatives & 0.0166 & 0.0652 & 0.0287  &0.0275&0.0231&0.0220&0.0297& 0.0351\\
Weak Positive & 0.0250 & 0.0250 &0.0409&0.0438&0.0278&0.0432&0.0297& 0.0369\\
Weak Neutral & 0.0206 & 0.0202 &0.0254&0.0218&0.0185&0.0174&0.0229& 0.0240\\
Strong Negative & 0.0101 & 0.0117 &0.0152&0.0133&0.0086&0.0076&0.0206& 0.0243\\
Strong Positive & 0.0199 & 0.0204&0.0189&0.0194&0.0084&0.0124&0.0557& 0.0531\\
Strong Neutral & 0.0074& 0.00756 &0.0096&0.0090&0.0044&0.0068&0.0125& 0.0157\\
\bottomrule
\end{tabular}
    
\end{table}

\paragraph{Personality:}
\begin{table}
\caption{Average personality score comparisons of human and machine generated data.}
\label{tab:personality}
\small
\begin{tabular}{ccccccccc}
     \toprule
 & \multicolumn{2}{|c|}{OpenAI} & \multicolumn{2}{|c|}{Wiki} & \multicolumn{2}{|c|}{Pubmed} & \multicolumn{2}{|c|}{Tweets} \\ 
 \cmidrule(r){1-9}
 \textbf{Metric}& \textbf{Human} & \textbf{Machine} & \textbf{Human} & \textbf{Machine} & \textbf{Human} & \textbf{Machine} & \textbf{Human} & \textbf{Machine}\\ 
\midrule 
Extroversion & 0.5667& 0.5156&0.1919&0.1921&0.3815&0.3643&0.4644& 0.4665\\
Neuroticism & 0.5188 & 0.5203 &0.1699&0.1692&0.6138&0.6343&0.64& 0.6432\\
Agreeableness & 0.6137& 0.3915 &0.2343&0.2353&0.4349&0.4265&0.5604& 0.566\\
Conscientiousness & 0.4517& 0.4649 &0.1987&0.1967&0.4425&0.4381&0.5561& 0.5628\\
Openness & 0.4238& 0.5921 &0.2052&0.2067&0.6179&0.0198&0.5019& 0.4983\\
\bottomrule
\end{tabular}
    
\end{table}

Research in psychology suggests that individual behavior and preferences can largely be explained by underlying personality traits. Traditionally, these traits are identified through surveys like the Big Five inventory questionnaire \cite{john1999big}, which asks participants to rate themselves on a 5-point scale across five traits: Extraversion, Agreeableness, Emotional Stability, Openness to experience, and Conscientiousness. Previous studies have shown that user-generated content, such as social media posts, can accurately predict personality \cite{Farnadi:2016}. To explore personality differences between machine-generated and user-generated content, we used the BERT personality prediction model\footnote{https://huggingface.co/Minej/bert-base-personality} to compute scores for the five personality traits.
Table \ref{tab:personality} presents the overall personality trait scores for the three datasets. The results indicate notable differences between machine-generated and human-generated texts. For instance, machine-generated texts are generally more open in the OpenAI dataset but significantly less open in the PubMed dataset. Scores for neuroticism and conscientiousness are quite similar between human and machine-generated texts across all datasets. These findings highlight the variability in personality scores depending on the source and nature of the text, indicating both alignments and significant differences between machine-generated and human-generated texts.

\section{Detection}
\label{result}
For the detection task, we defined the problem as a binary classification challenge. Specifically, the task is to determine whether a given input belongs to one of two categories. Let $F$ represent the vector in the feature space with $m$ features, $f_1, f_2,..., f_m$, and let $T$ represent the target vector in the output space with $n$ target categories $[0,1]$. The objective of the classification algorithm is to learn a model $M: F \rightarrow T$ that minimizes the prediction error over the blind test. In this paper, we explore the application of various machine learning algorithms for this detection task. Specifically, we compare eight algorithms: Logistic Regression, Random Forest, Multinomial NB, SGDClassifier\footnote{https://scikit-learn.org/stable/modules/generated/sklearn.linear\_model.SGDClassifier.html}, SVM, VotingClassifier\footnote{https://scikit-learn.org/stable/modules/generated/sklearn.ensemble.VotingClassifier.html} and
Sequential modeling\footnote{https://www.tensorflow.org/guide/keras/sequential\_model}.
For input features, we only used the content of the text. All algorithms were trained and tested using the same set of input features. In our experiments, each dataset was split into 90\% training data and 10\% test data. The training data was further used for 5-fold cross-validation (using 80\%-20\% split), and then the final model was tested on the 10\% test dataset. We report results for overall performance scores in Table \ref{tab:res_model} for the test datasets. For feature extraction, we used the TfidfVectorizer library from sklearn in Python\footnote{https://scikit-learn.org/stable/modules/generated/sklearn.feature\_extraction.text.TfidfVectorizer.html}. The performance outcomes of the algorithms for OpenAI, Wikipedia, Pubmed, and Twitter datasets are shown in  Table \ref{tab:res_openAI}, Table \ref{tab:res_abstract}, Table \ref{tab:res_wiki} and Table \ref{tab:res_tweet} respectively.

\begin{table}[ht]
\caption{Classification results (5-fold) on the four datasets. The reported results here are on the blind test sets. Here, Acc = Accuracy, P= Precision and R= Recall scores. The models: LR = Logistic Regression, DT = Decision Trees, RF = Random Forest, MNB = multinomial naive bayes, SGD = Stochastic Gradient Descent classifier, SVM = Support Vector Machine, VC = Majority Voting Classifier, and Seq = Sequential.}
    \label{tab:res_model}
    \small
\begin{minipage}[b]{0.45\linewidth}
\centering
\caption{OpenAI Dataset (xl-1542M)}
    \label{tab:res_openAI}
 \begin{tabular}{ccccc}
     \toprule
     \textbf{Model} &	\textbf{Acc} &	\textbf{P} &	\textbf{R} &	\textbf{F1} \\ 
     \midrule
LR	& 74.32\% &	75.89\% & 71.28\% & 73.51\% \\ 
DT & 58.99\% & 60.21\% & 61.63\% & 61.63\% \\ 
RF & 67.09\% & 71.56\% & 63.98\% & 67.56\%\\ 
MNB & 64.65\% &	61.57\% & 77.89\% &	68.81\% \\ 
SGD  & 64.88\% & 69.99\% & 52.07\% & 59.71\% \\ 
SVM	& 74.86\% &	73.89\% & 71.28\% &	72.56\% \\ 
VC & 72.69\% & 75.04\% & 70.53\% & 72.72\% \\ 
Seq & 73.60\% &	74.20\% & 71.05\% &	72.59\% \\ 
\bottomrule
\end{tabular}
\end{minipage}
\hspace{0.5cm}
\vspace{0.5cm}
\begin{minipage}[b]{0.45\linewidth}
\centering
 \caption{Wiki Dataset}
    \label{tab:res_wiki}
 \begin{tabular}{ccccc}
    \toprule
   \textbf{Model} &	\textbf{Acc} &	\textbf{P} &	\textbf{R} &	\textbf{F1} \\ 
   \midrule
    LR & 87.00\% & 86.81\% & 87.61\% & 87.21\% \\ 
    DT & 77.46\% & 77.08\% & 78.11\% & 77.59\% \\ 
    RF & 88.83\% & 88.92\% & 88.83\% & 88.82\% \\ 
    MNB & 74.58\% & 79.02\% & 66.87\% & 72.44\% \\ 
    SGD & 85.75\% & 85.28\% & 86.40\% & 85.83\% \\ 
    SVM & 92.28\% & 92.28\% & 92.28\% & 92.28\% \\ 
    VC & 88.27\% & 88.30\% & 88.27\% & 88.27\% \\ 
    Seq & 92.61\% & 92.05\% & 93.26\% & 92.65\% \\ 
\bottomrule
  \end{tabular}  
\end{minipage}
\begin{minipage}[b]{0.45\linewidth}
\centering
\caption{Abstract dataset}
    \label{tab:res_abstract}
  \begin{tabular}{ccccc}
    \toprule
   \textbf{Model} &	\textbf{Acc} &	\textbf{P} &	\textbf{R} &	\textbf{F1} \\ 
   \midrule
LR & 96.08\% &	96.10\% & 96.08\% & 96.08\%\\  
DT & 96.08\% &	96.10\% & 96.08\% &	96.08\% \\ 
RF & 97.32\% & 97.43\% & 97.32\% & 97.32\% \\ 
MNB & 87.40\% & 89.45\% & 87.40\% &	87.24\% \\ 
SGD & 90.08\% & 90.29\% & 90.08\% &	90.07\% \\ 
SVM	& 97.35\% & 97.37\% & 97.35\% &	97.35\% \\
VC & 97.02\% & 97.10\% & 97.02\% & 97.02\% \\ 
Seq & 96.82\% &	95.50\% & 95.50\% &	96.00\% \\ 
\bottomrule
 \end{tabular}
\end{minipage}
\hspace{0.85cm}
\vspace{0.5cm}
\begin{minipage}[b]{0.45\linewidth}
\centering
 \caption{Twitter Dataset}
    \label{tab:res_tweet}
  \begin{tabular}{ccccc}
    \toprule
   \textbf{Model} &	\textbf{Acc} &	\textbf{P} &	\textbf{R} &	\textbf{F1} \\ 
   \midrule
LR &  98.47\%&	 98.47\%&	 98.47\%&	 98.47\%\\ 
DT &  95.82\%&	 95.83\%&	 95.83\%&	 95.82\%\\ 
RF &  98.92\%&	 98.91\%&	 98.92\%&	 98.92\%\\ 
MNB	&  97.05\%&	 97.07\%& 97.07\%&	 97.05\%\\ 
SGD &  52.25\%&	 75.76\%&	 51.52\%&	 36.96\%\\ 
SVM &  98.02\%&	 98.01\%&	 98.02\%& 98.02\%\\ 
VC &  99.03\%&	 99.03\%&	 99.03\%&	 99.03\%\\ 
Seq &  99.00\%&	 99.00\%&	 99.00\%&	 99.00\%\\ 
\bottomrule
 \end{tabular}
\end{minipage}
\end{table}

The results in Table \ref{tab:res_model} demonstrate that SVM and VC consistently achieve high performance across most datasets, particularly excelling in the Wiki and Abstract datasets. In contrast, Decision Tree (DT) models generally exhibited the lowest performance across all datasets. While Logistic Regression (LR) performed well, it was outperformed by more complex models such as SVM and VC, especially in terms of accuracy and F1 scores. For the OpenAI dataset, the results presented pertain to the XL-1542M version, the largest version with 1.5 billion parameters, evaluated with a temperature setting of 1 (no truncation). The baseline performance for this dataset using Logistic Regression is reported to be 74.31\% \cite{gpt2outputdataset}. Although the performance of Logistic Regression on this dataset is well-documented, our study sought to evaluate the performance of other well-known models as well. Comprehensive results for all data versions are provided in Appendix Table \ref{tab:openAI_AllResults}. In the Twitter dataset (merged US Election, Game of Thrown and FIFA tweets), all models except SGD exhibited very high performance, with RF, VC, and Seq models approaching near-perfect metrics. This clearly demonstrates the limitation of machine data generated using generic prompt. Therefore, we further modified the Twitter and Abstract datasets using a rephrasing technique to increase the difficulty of the detection task (discussed in the following section). Overall, the results in Table \ref{tab:res_model} indicate that while traditional models like Logistic Regression and Decision Trees can perform adequately, advanced models such as SVM and ensemble methods like Voting Classifier tend to deliver superior performance, particularly in more complex or varied datasets.

\paragraph{Re-Phrased Text Detection}
\label{rephrase}
Detecting short and rephrased text poses challenges, as noted in previous studies \cite{Islam:2023}. In our initial experiments with Twitter and abstract datasets, we employed OpenAI’s GPT-3.5-turbo model to generate machine data, aiming to emulate diverse human behaviors. However, stark disparities emerged between human and machine-generated tweets. While human tweets often lack context, contain informal language with occasional errors, and express varied tones, machine-generated tweets tend to be comprehensive, positive, grammatically flawless, and formal \cite{Tang:2024}. To address this, we refined our prompt engineering approach by incorporating actual human tweets, instructing the model to mimic not only content but also style, tone, and vocabulary usage. We tested two methods of rephrasing: (1) generate data without any threshold constraint of vocabulary from the human text to generate machine data.; (2) generate data using at least 60\% of the vocabulary from the human text to generate machine data. This led to newly generated tweets and abstracts closely resembling their human counterparts. We report results for the second in this paper as it was harder to detect. Results for the first method is provided in the Table \ref{tab:rephrased_res_model_part2} of Appendix Section. 
As expected, detecting rephrased data proved more challenging, resulting in performance drops observed in both datasets. For instance, logistic regression's performance dropped from 96\% to 77\% for the Twitter dataset and from 98\% to 91\% for the PubMed dataset. These findings confirm previous literature, highlighting the increased difficulty in detecting shorter and rephrased texts, particularly in tweets compared to abstracts. It should be noted that the Twitter dataset include tweets on three different topics (US election, FIFA and Game of Thrown). Significant drop in the performance can also be due to the diversity in the topics. Results on individual topic is provided in the Table \ref{tab:rephrased_LR} of the Appendix Section.

\begin{table}[ht]
\caption{Classification results on the Twitter and Pubmed re-phrased datasets, using 60\% overlap constraint.}
    \label{tab:rephrased_res_model}
    \small
\begin{minipage}[b]{0.45\linewidth}
\centering
\caption{Twitter Dataset}
    \label{tab:rephrased_tweet}
 \begin{tabular}{ccccc}
     \toprule
     \textbf{Model} &	\textbf{Acc} &	\textbf{P} &	\textbf{R} &	\textbf{F1} \\ 
     \midrule
LR	&  77.43\%&	 77.55\%&	 77.42\%&	 77.4\%\\ 
DT &  65.64\%&	 65.65\%&	 65.64\%&	 65.63\%\\ 
RF &  73.47\%&	 73.53\%&	 73.47\%&	 73.45\%\\ 
MNB &  72.82\%&	 72.94\%&	 72.82\%&	 72.78\%\\ 
SGD  &  56.4\%&	 75.12\%&	 56.33\%&	 46.33\%\\ 
SVM	&  76.07\%&	 76.19\%&	 76.06\%&	 76.04\%\\ 
VC &  77.01\%&	 77.07\%&	 77.00\%&	 76.99\%\\ 
Seq &  77.02\%&	 77.15\%&	 77.02\%&	 77.00\%\\ 
\bottomrule
\end{tabular}
\end{minipage}
\hspace{0.5cm}
\begin{minipage}[b]{0.45\linewidth}
\centering
 \caption{Pubmed Dataset}
    \label{tab:rephrased_pubmed}
  \begin{tabular}{ccccc}
    \toprule
   \textbf{Model} &	\textbf{Acc} &	\textbf{P} &	\textbf{R} &	\textbf{F1} \\ 
   \midrule
LR & 91.22\% & 91.24\% & 91.22\% & 91.22\%	 \\ 
DT & 91.22\% & 91.24\% & 91.22\% & 91.22\% \\ 
RF & 89.86\% & 90.04\% & 89.86\% & 89.83\% \\ 
MNB	& 83.46\% &	84.00\% & 83.46\% &	83.34\% \\ 
SGD & 84.16\% &	84.18\% & 84.16\% &	84.16\% \\ 
SVM & 91.61\% &	91.62\% & 91.61\% &	91.61\% \\ 
VC & 84.72\% & 84.79\% & 84.72\% & 84.69\% \\ 
Seq & 82.24\% &	82.20\% & 82.20\% &	82.00\% \\ 
\bottomrule
 \end{tabular}
\end{minipage}
\end{table}

\section{Conclusion}
\label{conc}
In conclusion, this study contributes to the ongoing efforts to understand and effectively detect texts generated by large language models (LLMs). By analyzing datasets of varying text lengths and employing different machine learning algorithms, we gained insights into the performance and characteristics of human and machine-generated text. Our findings indicate that machine-generated texts generally exhibit higher readability and closely mimic human moral judgments, although notable differences in personality metrics were observed across datasets. Furthermore, SVM and VC consistently outperformed other algorithms, highlighting their effectiveness in detecting machine-generated text. However, our analysis is limited by the specific datasets and LLM versions used, underscoring the need for further research to generalize these findings and explore additional dimensions of human writing style. Moving forward, future studies will aim to address these limitations and enhance our understanding of LLM-generated texts, ultimately contributing to the development of more robust detection methods.

\bibliographystyle{plain}
\bibliography{bibfile}

\begin{thebibliography}{10}

\bibitem{aaditya_bhat_2023}
{Aaditya Bhat}.
\newblock Gpt-wiki-intro (revision 0e458f5), 2023.

\bibitem{abdali:2024}
Sara Abdali, Richard Anarfi, CJ~Barberan, and Jia He.
\newblock Decoding the ai pen: Techniques and challenges in detecting ai-generated text, 2024.

\bibitem{Abdalla:2023}
Mohamed Hesham~Ibrahim Abdalla, Simon Malberg, Daryna Dementieva, Edoardo Mosca, and Georg Groh.
\newblock A benchmark dataset to distinguish human-written and machine-generated scientific papers.
\newblock {\em Information}, 14(10):522, 2023.

\bibitem{arnold:2022}
M.~Arnold, J.~Yang, and J.~May.
\newblock Detecting computer-generated text using stylometric features.
\newblock {\em Journal of Artificial Intelligence Research}, 65:923--947, 2022.

\bibitem{beguš:2023}
Gašper Beguš, Maksymilian Dąbkowski, and Ryan Rhodes.
\newblock Large linguistic models: Analyzing theoretical linguistic abilities of llms, 2023.

\bibitem{bender:2021}
E.~M. Bender, T.~Gebru, A.~McMillan-Major, and M.~Mitchell.
\newblock On the dangers of stochastic parrots: Can language models be too big?
\newblock {\em arXiv preprint}, 2021.

\bibitem{Chong:2023}
A.~T.~Y. Chong, H.~N. Chua, M.~B. Jasser, and R.~T.~K. Wong.
\newblock Bot or human? detection of deepfake text with semantic, emoji, sentiment and linguistic features.
\newblock In {\em 2023 IEEE 13th International Conference on System Engineering and Technology (ICSET)}, pages 205--210, Shah Alam, Malaysia, 2023.

\bibitem{deshpande:2023}
Ameet Deshpande, Vishvak Murahari, Tanmay Rajpurohit, Ashwin Kalyan, and Karthik Narasimhan.
\newblock Toxicity in chatgpt: Analyzing persona-assigned language models.
\newblock In Houda Bouamor, Juan Pino, and Kalika Bali, editors, {\em Findings of the Association for Computational Linguistics: EMNLP 2023}, pages 1236--1270, Singapore, December 2023. Association for Computational Linguistics.

\bibitem{Duarte:2024}
Fabio Duarte.
\newblock Number of chatgpt users, January 2024.

\bibitem{Else:2023}
Holly Else.
\newblock {Abstracts written by ChatGPT fool scientists}.
\newblock {\em Nature}, 613(7944):423--423, January 2023.

\bibitem{Rooney:2023}
Michael Elsen-Rooney.
\newblock Nyc education department blocks chatgpt on school devices, networks, January 2023.

\bibitem{Emsley:2023}
Robin~A Emsley.
\newblock Chatgpt: these are not hallucinations – they’re fabrications and falsifications.
\newblock {\em Schizophrenia}, 9, 2023.

\bibitem{Farnadi:2016}
Golnoosh Farnadi, Geetha Sitaraman, Shanu Sushmita, Fabio Celli, Michal Kosinski, David Stillwell, Sergio Davalos, Marie-Francine Moens, and Martine Cock.
\newblock Computational personality recognition in social media.
\newblock 26(2–3), 2016.

\bibitem{Gehrmann:2019}
Sebastian Gehrmann, Hendrik Strobelt, and Alexander~M. Rush.
\newblock Gltr: Statistical detection and visualization of generated text.
\newblock {\em CoRR}, abs/1906.04043, 2019.

\bibitem{graham:2013}
Jesse Graham, Jonathan Haidt, Sena Koleva, Matt Motyl, Ravi Iyer, Sean~P. Wojcik, and Peter~H. Ditto.
\newblock Moral foundations theory: The pragmatic validity of moral pluralism.
\newblock In {\em Advances in Experimental Social Psychology}, volume~47, pages 55--130. Elsevier, 2013.

\bibitem{guo2023}
Biyang Guo, Xin Zhang, Ziyuan Wang, Minqi Jiang, Jinran Nie, Yuxuan Ding, Jianwei Yue, and Yupeng Wu.
\newblock How close is chatgpt to human experts? comparison corpus, evaluation, and detection, 2023.

\bibitem{gururangan:2020}
Suchin Gururangan, Ana Marasović, Swabha Swayamdipta, Kyle Lo, Iz~Beltagy, Doug Downey, and Noah~A. Smith.
\newblock Don’t stop pretraining: Adapt language models to domains and tasks.
\newblock In {\em Proceedings of ACL}, 2020.

\bibitem{horne2019robust}
Benjamin~D Horne, Jeppe N{\o}rregaard, and Sibel Adali.
\newblock Robust fake news detection over time and attack.
\newblock {\em ACM Transactions on Intelligent Systems and Technology (TIST)}, 11(1):1--23, 2019.

\bibitem{Islam:2023}
Niful Islam, Debopom Sutradhar, Humaira Noor, Jarin~Tasnim Raya, Monowara~Tabassum Maisha, and Dewan~Md Farid.
\newblock Distinguishing human generated text from chatgpt generated text using machine.
\newblock {\em arXiv}, 2023.

\bibitem{Jawahar:2020}
Ganesh Jawahar, Muhammad Abdul-Mageed, and Laks V.~S. Lakshmanan.
\newblock Automatic detection of machine generated text: A critical survey.
\newblock In {\em International Conference on Computational Linguistics}, 2020.

\bibitem{john1999big}
Oliver~P John and Sanjay Srivastava.
\newblock The {B}ig {F}ive trait taxonomy: History, measurement, and theoretical perspectives.
\newblock {\em Handbook of Personality: Theory and Research}, 2:102--138, 1999.

\bibitem{li:2023}
Cheng Li, Jindong Wang, Yixuan Zhang, Kaijie Zhu, Wenxin Hou, Jianxun Lian, Fang Luo, Qiang Yang, and Xing Xie.
\newblock Large language models understand and can be enhanced by emotional stimuli, 2023.

\bibitem{liao2023differentiate}
Wenxiong Liao, Zhengliang Liu, Haixing Dai, Shaochen Xu, Zihao Wu, Yiyang Zhang, Xiaoke Huang, Dajiang Zhu, Hongmin Cai, Tianming Liu, and Xiang Li.
\newblock Differentiate chatgpt-generated and human-written medical texts, 2023.

\bibitem{Liu:2005}
Bing Liu, Minqing Hu, and Junsheng Cheng.
\newblock Opinion observer: analyzing and comparing opinions on the web.
\newblock In {\em Proceedings of the 14th International Conference on World Wide Web}, WWW '05, page 342–351, New York, NY, USA, 2005. Association for Computing Machinery.

\bibitem{McCLURE:1987}
Glenda~M. McCLURE.
\newblock Readability formulas: Useful or useless?
\newblock {\em IEEE Transactions on Professional Communication}, PC-30:12--15, 1987.

\bibitem{mitrović2023chatgpt}
Sandra Mitrović, Davide Andreoletti, and Omran Ayoub.
\newblock Chatgpt or human? detect and explain. explaining decisions of machine learning model for detecting short chatgpt-generated text, 2023.

\bibitem{gpt2outputdataset}
OpenAI.
\newblock Gpt-2 output dataset.
\newblock \url{https://github.com/openai/gpt-2-output-dataset}, 2019.
\newblock Accessed on: 2024.

\bibitem{Oravec:2023}
Jo~Ann Oravec.
\newblock Artificial intelligence implications for academic cheating: Expanding the dimensions of responsible human-ai collaboration with chatgpt and bard.
\newblock 2023.

\bibitem{Pan:2023}
Jonathan Pan.
\newblock A foundation model approach to detect machine generated text.
\newblock volume~54, pages 405--408, Chiang Mai, Thailand, Apr 2023.

\bibitem{radford:2018}
Alec Radford, Karthik Narasimhan, Tim Salimans, and Ilya Sutskever.
\newblock Improving language understanding by generative pretraining, 2018.
\newblock OpenAI Blog.

\bibitem{sap:2020}
M.~Sap, D.~Card, S.~Gabriel, Y.~Choi, N.~A. Smith, and Y.~Choi.
\newblock The risk of racial bias in hate speech detection.
\newblock In {\em Proceedings of the AAAI/ACM Conference on AI, Ethics, and Society}, 2020.

\bibitem{solaiman:2019}
Irene Solaiman, Miles Brundage, Jack Clark, Amanda Askell, Ariel Herbert-Voss, Jeff Wu, Alec Radford, and Jasmine Wang.
\newblock Release strategies and the social impacts of language models.
\newblock {\em ArXiv}, abs/1908.09203, 2019.

\bibitem{Taloni:2023}
Andrea Taloni, Vincenzo Scorcia, and Giuseppe Giannaccare.
\newblock {Large Language Model Advanced Data Analysis Abuse to Create a Fake Data Set in Medical Research}.
\newblock {\em JAMA Ophthalmology}, 141(12):1174--1175, 12 2023.

\bibitem{tang2023}
Ruixiang Tang, Yu-Neng Chuang, and Xia Hu.
\newblock The science of detecting llm-generated texts, 2023.

\bibitem{Tang:2024}
Ruixiang Tang, Yu-Neng Chuang, and Xia Hu.
\newblock The science of detecting llm-generated text.
\newblock {\em Commun. ACM}, 67(4):50–59, mar 2024.

\bibitem{Lancet:2024}
{The Lancet Digital Health}.
\newblock Large language models: a new chapter in digital health.
\newblock {\em The Lancet Digital Health}, 6(1):e1, 2024.

\bibitem{Varanasi:2023}
Lakshmi Varanasi.
\newblock Gpt-4 can ace the bar, but it only has a decent chance of passing the cfa exams. here's a list of difficult exams the chatgpt and gpt-4 have passed., Nov 2023.

\bibitem{xie:2024}
Yaqi Xie, Anjali Rawal, Yujing Cen, Dixuan Zhao, Sunil~K Narang, and Shanu Sushmita.
\newblock Mugc: Machine generated versus user generated content detection, 2024.

\bibitem{Yang:2023}
Xianjun Yang, Liangming Pan, Xuandong Zhao, Haifeng Chen, Linda~Ruth Petzold, William~Yang Wang, and Wei Cheng.
\newblock A survey on detection of llms-generated content.
\newblock {\em ArXiv}, abs/2310.15654, 2023.

\bibitem{yucheat:2023}
Peipeng Yu, Jiahan Chen, Xuan Feng, and Zhihua Xia.
\newblock Cheat: A large-scale dataset for detecting chatgpt-written abstracts, 2023.

\bibitem{Zellers:2019}
Rowan Zellers, Ari Holtzman, Hannah Rashkin, Yonatan Bisk, Ali Farhadi, Franziska Roesner, and Yejin Choi.
\newblock Defending against neural fake news.
\newblock {\em ArXiv}, abs/1905.12616, 2019.

\end{thebibliography}

\newpage
\appendix

\section{Appendix / supplemental material}

\begin{table}[ht]
   \caption{Datasets and their corresponding columns.}
    \label{tab:col_names}
    \small
    \begin{tabular}{cc}
     \toprule 
     \textbf{Dataset} & \textbf{Column Names} \\
     \midrule
 OpenAI GPT2 & id, text, length, ended, label \\ 
 Wiki Intro	&  id, title, wiki intro, generated intro, label\\
 Pubmed  & id, abstract text and label\\
 Twitter & id, tweets (full tweet text) and label \\
\bottomrule
 \end{tabular}
\end{table}

\begin{table}[h]
\caption{Readability score of human and machine generated data. Here standard deviation for all four datasets are reported. }
\label{tab:readSD}
\small
\begin{tabular}{ccccccccc}
     \toprule
 & \multicolumn{2}{|c|}{OpenAI} & \multicolumn{2}{|c|}{Wiki} & \multicolumn{2}{|c|}{Pubmed} & \multicolumn{2}{|c|}{Tweets} \\ 
 \cmidrule(r){1-9}
 \textbf{Metric}& \textbf{Human} & \textbf{Machine} & \textbf{Human} & \textbf{Machine} & \textbf{Human} & \textbf{Machine} & \textbf{Human} & \textbf{Machine}\\ 
\midrule 
\small{Gunning Fog Index} & 7.73 & 6.11 &5.09&2.42&2.21&3.56&6.45& 5.44\\
SMOG Index &3.17 & 3.37 &2.60& 3.01& 2.41&3.99&4.43& 3.96\\
Dale-Chall Readability & 1.77 & 1.92 &1.17&1.10&1.00&1.55&3.07& 2.64\\
Flesch Reading Ease Score & 24.77 & 88.69 &18.00 &13.24&11.84&16.33&27.87& 24.80\\
Coleman Liau Index & 4.49 & 18.72 & 2.34&2.23&3.56&3.33&9.05& 6.72\\
\bottomrule
\end{tabular}    
\end{table}

\begin{table}[h]
\caption{Bias metric comparisons between human and machine generated data. Here, standard deviation scores for datasets are reported. }
\label{tab:bias_SD}
\small
\begin{tabular}{ccccccccc}
     \toprule
 & \multicolumn{2}{|c|}{OpenAI} & \multicolumn{2}{|c|}{Wiki} & \multicolumn{2}{|c|}{Pubmed} & \multicolumn{2}{|c|}{Tweets} \\ 
 \cmidrule(r){1-9}
 \textbf{Metric}& \textbf{Human} & \textbf{Machine} & \textbf{Human} & \textbf{Machine} & \textbf{Human} & \textbf{Machine} & \textbf{Human} & \textbf{Machine}\\ 
\midrule 
Bias words & 0.0281 & 0.0265 &0.0703& 0.0874&0.020& 0.02&0.0842& 0.0865\\
Assertatives & 0.0055 & 0.0052 &0.0134& 0.0165&0.0041&0.0047&0.0217& 0.0234\\
Factives & 0.0043& 0.0043 & 0.0080& 0.0114& 0.006 &0.008&0.0169& 0.0196\\
Hedges & 0.0096 & 0.0090 & 0.0134& 0.0169& 0.008& 0.007&0.0220& 0.0242\\
Implicatives & 0.0069 & 0.0063 &0.0084 & 0.0111& 0.008&0.01&0.0233& 0.0303\\
\bottomrule
\end{tabular}    
\end{table}

\begin{table}
\small
\caption{Morality comparisons between human and machine generated data. Here, standard deviation scores for datasets are reported.}
\label{tab:moral_SD}
\begin{tabular}{ccccccccc}
     \toprule
 & \multicolumn{2}{|c|}{OpenAI} & \multicolumn{2}{|c|}{Wiki} & \multicolumn{2}{|c|}{Pubmed} & \multicolumn{2}{|c|}{Tweets} \\ 
 \cmidrule(r){1-9}
 \textbf{Metric}& \textbf{Human} & \textbf{Machine} & \textbf{Human} & \textbf{Machine} & \textbf{Human} & \textbf{Machine} & \textbf{Human} & \textbf{Machine}\\ 
\midrule 
 Harm & 0.0046  & 0.0039 & 0.0091& 0.0131& 0.003 & 0.004&0.0096& 0.0102\\
Fairness & 0.0013 & 0.0014 & 0.0024 & 0.0027& 0.0008& 0.0005&0.0042& 0.0047\\
Cheating & 0.0005 & 0.0005 & 0.0091& 0.0131&0.0012&0.0006&0.0014& 0.0016\\
Loyalty & 0.0038& 0.0035 & 0.0106&0.0132&0.005&0.002&0.0077& 0.0067\\
Betrayal & 0.0011 & 0.0010& 0.0023& 0.0029 &0.0005&.0002&0.0039& 0.0031\\
Authority & 0.0049 & 0.0044 & 0.0117 & 0.0129&0.0044&.0014&0.0083& 0.0080\\
Subversion & 0.0015 & 0.0012 & 0.0025& 0.0036&0.0006&0.0006&0.0045& 0.0038\\
Purity & 0.0019 & 0.0017 &0.0082&0.0091&0.0005&.0006&0.0032& 0.0038\\
Degradation & 0.0013 & 0.0012 &0.0024&0.0031&0.0007&0.0005&0.0063& 0.0082\\
Morality General & 0.0033 & 0.0034 &0.0048&0.0065&0.005&0.001&0.0161& 0.0140\\
\bottomrule
\end{tabular}    
\end{table}

\begin{table}
\caption{Sentiment comparisons between human and machine generated data. Here, standard deviation scores for datasets are reported. }
\label{tab:sentiment}
\small
\begin{tabular}{ccccccccc}
     \toprule
 & \multicolumn{2}{|c|}{OpenAI} & \multicolumn{2}{|c|}{Wiki} & \multicolumn{2}{|c|}{Pubmed} & \multicolumn{2}{|c|}{Tweets} \\ 
 \cmidrule(r){1-9}
 \textbf{Metric}& \textbf{Human} & \textbf{Machine} & \textbf{Human} & \textbf{Machine} & \textbf{Human} & \textbf{Machine} & \textbf{Human} & \textbf{Machine}\\ 
\midrule 
Weak Negatives & 0.0129 & 0.0119&0.0289&0.0369&0.0179&0.193&0.0395& 0.0449\\
Weak Positive & 0.0151& 0.0142&0.0277&0.0378&0.0191&0.0228&0.0424& 0.0467\\
Weak Neutral &0.0131& 0.0124&0.0206&0.0244&0.0185&0.0174&0.0339& 0.0384\\
Strong Negative & 0.0102 & 0.0100&0.0192&0.0230&0.0866&0.0766&0.0373& 0.0399\\
Strong Positive &0.0150& 0.0142&0.0207&0.0265&0.0099&0.0118&0.0877& 0.0655\\
Strong Neutral &0.0078&0.0075&0.0119&0.0149&0.0066&0.0068&0.0261& 0.0309\\
\bottomrule
\end{tabular}    
\end{table}

\begin{table}
\caption{Standard Deviation personality score comparisons of human and machine generated data.}
\label{tab:personality_SD}
\small
\begin{tabular}{ccccccccc}
     \toprule
 & \multicolumn{2}{|c|}{OpenAI} & \multicolumn{2}{|c|}{Wiki} & \multicolumn{2}{|c|}{Pubmed} & \multicolumn{2}{|c|}{Tweets} \\ 
 \cmidrule(r){1-9}
 \textbf{Metric}& \textbf{Human} & \textbf{Machine} & \textbf{Human} & \textbf{Machine} & \textbf{Human} & \textbf{Machine} & \textbf{Human} & \textbf{Machine}\\ 
\midrule 
Extroversion &0.0243 & 0.0241&0.0054&0.0060&0.0553&0.0175&0.0216& 0.0229\\
Neuroticism & 0.0226&0.0157&0.0055&0.0061&0.0666&0.0173&0.0128& 0.0121\\
Agreeableness & 0.0153&0.0219&0.0067&0.0074&0.0476&0.0331&0.0234& 0.0223\\
Conscientiousness & 0.0209&0.0373&0.0065&0.0067&0.0472&0.0266&0.028& 0.0269\\
Openness & 0.0260&0.0243&0.0066&0.0073&0.0198&.0154&0.0142& 0.0142\\
\bottomrule
\end{tabular}    
\end{table}

\begin{table}[ht]
\centering
\caption{Model Performance on Train and Test Sets for complete Open AI Dataset}
\label{tab:openAI_AllResults}
\small
\begin{tabular}{lcccccccc}
\toprule
& \multicolumn{4}{c}{Dev Set} & \multicolumn{4}{c}{Test Set} \\
\cmidrule(r){2-5} \cmidrule(r){6-9}
Model & Accuracy & Precision & Recall & F1 & Accuracy & Precision & Recall & F1 \\
\midrule
\multicolumn{9}{c}{large-762M} \\
lr & 76.66\% & 78.76\% & 73.02\% & 75.78\% & 77.28\% & 78.28\% & 73.86\% & 76.01\% \\
sgd & 67.67\% & 73.29\% & 55.60\% & 63.23\% & 67.75\% & 73.24\% & 55.94\% & 63.43\% \\
nb & 64.78\% & 60.86\% & 82.82\% & 70.16\% & 65.89\% & 62.32\% & 80.40\% & 70.21\% \\
dt & 61.17\% & 61.26\% & 60.73\% & 60.99\% & 60.42\% & 60.37\% & 60.64\% & 60.50\% \\
rf & 69.25\% & 73.73\% & 59.80\% & 66.04\% & 69.45\% & 73.86\% & 60.22\% & 66.35\% \\
svm(linear) & 78.34\% & 78.41\% & 73.93\% & 76.10\% & 78.42\% & 78.33\% & 73.74\% & 75.97\% \\
svm(sigmoid) & 75.47\% & 77.69\% & 74.05\% & 75.83\% & 75.61\% & 77.56\% & 72.58\% & 74.99\% \\
vc & 76.17\% & 76.81\% & 72.39\% & 74.53\% & 75.83\% & 77.17\% & 70.50\% & 73.68\% \\
MLP & 71.33\% & 74.20\% & 73.52\% & 73.86\% & 71.61\% & 73.68\% & 74.84\% & 74.26\% \\
\midrule
\multicolumn{9}{c}{large-762M-k40} \\
lr & 93.82\% & 93.79\% & 93.84\% & 93.81\% & 94.39\% & 94.74\% & 94.00\% & 94.37\% \\
sgd & 85.06\% & 88.75\% & 80.28\% & 84.30\% & 84.81\% & 88.31\% & 80.24\% & 84.08\% \\
nb & 77.51\% & 72.31\% & 89.14\% & 79.85\% & 77.38\% & 72.46\% & 88.34\% & 79.62\% \\
dt & 63.84\% & 65.40\% & 62.32\% & 63.82\% & 64.21\% & 64.79\% & 63.43\% & 64.10\% \\
rf & 85.47\% & 87.49\% & 85.71\% & 86.59\% & 85.60\% & 87.53\% & 84.88\% & 86.18\% \\
svm(linear) & 94.46\% & 94.89\% & 92.32\% & 93.59\% & 93.98\% & 94.82\% & 91.75\% & 93.26\% \\
svm(sigmoid) & 93.36\% & 92.05\% & 90.70\% & 91.37\% & 93.37\% & 90.30\% & 92.46\% & 91.37\% \\
vc & 94.49\% & 94.76\% & 94.98\% & 94.87\% & 93.48\% & 96.19\% & 95.97\% & 96.08\% \\
MLP & 88.29\% & 87.91\% & 88.25\% & 88.08\% & 87.94\% & 88.34\% & 87.08\% & 87.71\% \\
\midrule
\multicolumn{9}{c}{medium-345M} \\
lr & 88.69\% & 90.58\% & 86.36\% & 88.42\% & 89.07\% & 91.01\% & 86.70\% & 88.80\% \\
sgd & 82.61\% & 91.28\% & 72.13\% & 80.58\% & 82.64\% & 91.19\% & 72.26\% & 80.63\% \\
nb & 80.96\% & 79.68\% & 83.11\% & 81.36\% & 81.56\% & 81.21\% & 82.12\% & 81.66\% \\
dt & 68.61\% & 69.92\% & 63.85\% & 66.75\% & 68.72\% & 70.38\% & 64.73\% & 67.44\% \\
rf & 85.86\% & 92.30\% & 85.55\% & 88.80\% & 84.23\% & 89.40\% & 86.51\% & 87.93\% \\
svm(linear) & 88.44\% & 89.64\% & 87.45\% & 88.53\% & 88.16\% & 90.63\% & 87.44\% & 89.01\% \\
svm(sigmoid) & 85.25\% & 88.74\% & 87.93\% & 88.33\% & 86.08\% & 88.36\% & 87.58\% & 87.97\% \\
vc & 89.17\% & 91.12\% & 85.30\% & 88.11\% & 90.79\% & 90.56\% & 84.95\% & 87.67\% \\
MLP & 84.86\% & 88.65\% & 86.29\% & 87.45\% & 84.04\% & 87.52\% & 85.07\% & 86.28\% \\
\midrule
\multicolumn{9}{c}{medium-345M-k40} \\
lr & 94.77\% & 94.93\% & 94.58\% & 94.75\% & 95.26\% & 95.57\% & 94.92\% & 95.24\% \\
sgd & 84.78\% & 89.27\% & 79.05\% & 83.85\% & 84.32\% & 88.59\% & 78.78\% & 83.40\% \\
nb & 78.47\% & 72.57\% & 91.55\% & 80.96\% & 78.28\% & 72.85\% & 90.16\% & 80.59\% \\
dt & 76.63\% & 76.81\% & 72.55\% & 74.62\% &

 75.46\% & 76.47\% & 72.37\% & 74.36\% \\
rf & 88.39\% & 90.55\% & 89.34\% & 89.94\% & 89.27\% & 90.59\% & 88.83\% & 89.70\% \\
svm(linear) & 93.68\% & 94.82\% & 91.69\% & 93.23\% & 93.87\% & 94.34\% & 90.47\% & 92.36\% \\
svm(sigmoid) & 92.96\% & 91.52\% & 90.67\% & 91.09\% & 93.43\% & 91.09\% & 92.41\% & 91.74\% \\
vc & 95.32\% & 95.57\% & 95.73\% & 95.65\% & 94.79\% & 95.89\% & 96.13\% & 96.01\% \\
MLP & 90.59\% & 90.10\% & 90.79\% & 90.44\% & 89.95\% & 90.76\% & 89.78\% & 90.27\% \\
\midrule
\multicolumn{9}{c}{small-117M} \\
lr & 83.92\% & 86.97\% & 79.58\% & 83.11\% & 83.41\% & 86.59\% & 78.58\% & 82.41\% \\
sgd & 78.64\% & 82.43\% & 73.41\% & 77.65\% & 78.51\% & 82.33\% & 73.24\% & 77.55\% \\
nb & 73.19\% & 69.91\% & 81.99\% & 75.47\% & 72.97\% & 69.78\% & 81.68\% & 75.30\% \\
dt & 63.56\% & 64.08\% & 62.78\% & 63.42\% & 63.78\% & 64.39\% & 62.87\% & 63.62\% \\
rf & 82.67\% & 85.29\% & 80.82\% & 83.00\% & 82.96\% & 85.71\% & 80.81\% & 83.17\% \\
svm(linear) & 84.53\% & 88.29\% & 81.10\% & 84.54\% & 84.94\% & 88.60\% & 81.54\% & 84.91\% \\
svm(sigmoid) & 83.69\% & 84.92\% & 83.11\% & 84.01\% & 83.71\% & 84.76\% & 83.26\% & 84.00\% \\
vc & 84.87\% & 87.67\% & 85.56\% & 86.60\% & 84.96\% & 87.77\% & 85.75\% & 86.75\% \\
MLP & 79.94\% & 80.82\% & 81.21\% & 81.02\% & 79.82\% & 80.68\% & 81.15\% & 80.92\% \\
\midrule
\multicolumn{9}{c}{small-117M-k40} \\
lr & 89.53\% & 91.12\% & 88.05\% & 89.55\% & 90.11\% & 91.93\% & 88.12\% & 89.99\% \\
sgd & 82.53\% & 89.52\% & 77.16\% & 82.81\% & 82.61\% & 89.38\% & 77.12\% & 82.74\% \\
nb & 74.19\% & 70.48\% & 82.95\% & 76.18\% & 73.87\% & 70.18\% & 82.68\% & 75.96\% \\
dt & 68.19\% & 68.81\% & 65.62\% & 67.18\% & 68.41\% & 69.06\% & 66.03\% & 67.50\% \\
rf & 82.46\% & 86.83\% & 83.79\% & 85.28\% & 82.61\% & 86.95\% & 83.93\% & 85.41\% \\
svm(linear) & 88.38\% & 89.61\% & 87.28\% & 88.43\% & 88.21\% & 89.49\% & 87.05\% & 88.26\% \\
svm(sigmoid) & 87.71\% & 88.40\% & 86.57\% & 87.48\% & 87.94\% & 88.50\% & 86.76\% & 87.63\% \\
vc & 89.58\% & 90.85\% & 89.77\% & 90.31\% & 89.43\% & 90.82\% & 89.57\% & 90.19\% \\
MLP & 84.36\% & 85.46\% & 84.08\% & 84.76\% & 84.31\% & 85.55\% & 84.12\% & 84.83\% \\
\bottomrule
\end{tabular}
\end{table}

\begin{table}[]
\caption{Classification results on the Twitter and Pubmed re-phrased datasets using any percentage of the vocabulary from the human text to generate machine data}
    \label{tab:rephrased_res_model_part2}
    \small
\begin{minipage}[b]{0.45\linewidth}
\centering
\caption{Twitter Dataset}
    \label{tab:rephrased_tweet_part2}
 \begin{tabular}{ccccc}
     \toprule
     \textbf{Model} &	\textbf{Acc} &	\textbf{P} &	\textbf{R} &	\textbf{F1} \\ 
     \midrule
LR	&  82.97\%&  82.65\%&	 78.02\%&	  79.58\%\\  
DT &  74.53\%&  71.42\%&	 71.41\%&	 	 71.42\%\\  
RF &  82.10\%&  82.59\%&	 76.14\%&	 	 78.00\%\\ 
MNB & 72.62\%&  80.87\%&	 59.67\%&	 	 58.05\%\\  
SGD  &  66.50\%&  33.25\%&	 50.00\%&	 	 39.94\%\\ 
SVM	&  80.60\%&  78.21\%&	 79.49\%&	 	 78.74\%\\ 
VC &  83.37\%&  84.16\%&	 77.70\%&	 	 79.63\%\\ 
Seq &  83.24\%&  82.99\%&	 83.24\%&	 	 82.79\%\\ 
\bottomrule
\end{tabular}
\end{minipage}
\hspace{0.5cm}
\begin{minipage}[b]{0.45\linewidth}
\centering
 \caption{Pubmed Dataset}
    \label{tab:rephrased_pubmed_part2}
  \begin{tabular}{ccccc}
    \toprule
   \textbf{Model} &	\textbf{Acc} &	\textbf{P} &	\textbf{R} &	\textbf{F1} \\ 
   \midrule
LR & 92.76\% & 93.40\% & 92.76\% & 92.53\% \\ 
DT & 92.76\% & 93.4\% &	92.76\% & 92.53\% \\ 
RF & 95.24\% & 95.27\% & 95.24\% & 95.20\% \\ 
MNB	& 90.67\% &	91.36\% & 90.67\% &	90.33\% \\ 
SGD & 91.65\% &	91.63\% & 91.65\% &	91.58\% \\ 
SVM & 95.92\% &	95.98\% & 95.92\% &	95.88\% \\ 
VC & 93.15\% & 93.28\% & 93.15\% & 93.04\% \\ 
Seq & 91.15\% &	91.50\% & 91.50\% &	91.00\% \\ 
\bottomrule
 \end{tabular}
\end{minipage}
\end{table}

\begin{table}[]
\caption{Classification results for Logistic Regression (best performing model) on the rephrased Tweets using 60\% overlap constraint.}
    \label{tab:rephrased_LR}
\small
\begin{minipage}[b]{0.45\linewidth}
\centering
\caption{Train}
    \label{tab:train_fifa}
 \begin{tabular}{ccccc}
     \toprule
     \textbf{Topic} &	\textbf{Acc} &	\textbf{P} &	\textbf{R} &	\textbf{F1} \\ 
     \midrule
FIFA	& 76.56\%& 76.62\%&76.53\%&	76.53\%\\
Election &	72.51\%&72.78\%& 72.49\%&	72.42\%\\
Game of Thrones	& 83.71\%&	83.71\%& 83.71\%&	83.7\%\\
Merged (all three) &	77.62\%& 77.75\%&	77.6\%& 77.58\%\\
\bottomrule
\end{tabular}
\end{minipage}
\hspace{2 cm}
\begin{minipage}[b]{0.45\linewidth}
\centering
 \caption{Test}
    \label{tab:test_fifa}
  \begin{tabular}{cccc}
    \toprule
   	\textbf{Acc} &	\textbf{P} &	\textbf{R} &	\textbf{F1} \\ 
   \midrule
78.15\%&78.48\%&	78.11\%& 78.07\%\\
71.31\%& 71.57\%& 71.18\%&	71.13\%\\
82.2\%& 82.2\%&	82.18\%&	82.19\%\\
77.43\%& 77.55\%&	77.42\%& 77.4\%\\
\bottomrule
 \end{tabular}
\end{minipage}
\end{table}


\end{document}